# Language Models Explain Word Reading Times Better Than Empirical Predictability

*Markus J. Hofmann[1]\*, Steffen Remus[2], Chris Biemann[2], Ralph Radach[1] and Lars Kuchinke[3]*

[1] *Department of Psychology, University of Wuppertal, Wuppertal, Germany,* [2] *Department of Informatics, Universität Hamburg, Hamburg, Germany,* [3] *International Psychoanalytic University, Berlin, Germany*

Though there is a strong consensus that word length and frequency are the most important single-word features determining visual-orthographic access to the mental lexicon, there is less agreement as how to best capture syntactic and semantic factors. The traditional approach in cognitive reading research assumes that word predictability from sentence context is best captured by cloze completion probability (CCP) derived from human performance data. We review recent research suggesting that probabilistic language models provide deeper explanations for syntactic and semantic effects than CCP. Then we compare CCP with three probabilistic language models for predicting word viewing times in an English and a German eye tracking sample: (1) Symbolic n-gram models consolidate syntactic and semantic short-range relations by computing the probability of a word to occur, given two preceding words. (2) Topic models rely on subsymbolic representations to capture long-range semantic similarity by word co-occurrence counts in documents. (3) In recurrent neural networks (RNNs), the subsymbolic units are trained to predict the next word, given all preceding words in the sentences. To examine lexical retrieval, these models were used to predict single fixation durations and gaze durations to capture rapidly successful and standard lexical access, and total viewing time to capture late semantic integration. The linear item-level analyses showed greater correlations of all language models with all eye-movement measures than CCP. Then we examined non-linear relations between the different types of predictability and the reading times using generalized additive models. N-gram and RNN probabilities of the present word more consistently predicted reading performance compared with topic models or CCP. For the effects of last-word probability on current-word viewing times, we obtained the best results with n-gram models. Such count-based models seem to best capture short-range access that is still underway when the eyes move on to the subsequent word. The prediction-trained RNN models, in contrast, better predicted early preprocessing of the next word. In sum, our results demonstrate that the different language models account for differential cognitive processes during reading. We discuss these algorithmically concrete blueprints of lexical consolidation as theoretically deep explanations for human reading.

**Keywords: language models, n-gram model, topic model, recurrent neural network model, predictability, generalized additive models, eye movements**

# INTRODUCTION

Concerning the influence of single-word properties, there is a strong consensus in the word recognition literature that word length and frequency are the most reliable predictors of lexical access (e.g., Reichle et al., 2003; New et al., 2006; Adelman and Brown, 2008; Brysbaert et al., 2011). Though for instance, Baayen (2010) suggests that a large part of the variance explained by word frequency is better explained by contextual word features, we here use these single-word properties as a baseline to set the challenge for contextual word properties to explain more variance than the single-word properties.

In contrast to single-word frequency, the question of how to best capture contextual word properties is controversial. The traditional psychological predictor variables are based on human performance. When aiming to quantify how syntactic and semantic contextual word features influence the reading of the present word, Taylor's (1953) cloze completion probability (CCP) still represents the performance-based state of the art for predicting sentence reading in psychological research (Kutas and Federmeier, 2011; Staub, 2015). Participants of a pre-experimental study are given a sentence with a missing word, and the relative number of participants completing the respective word are then taken to define CCP. This human performance is then used to account for another human performance such as reading. Westbury (2016), however, suggests that a to-be-explained variable, the explanandum, should be selected from a different domain than the explaining variable, the explanans (Hempel and Oppenheim, 1948). When two directly observable variables, such as CCP and reading times, are connected, for instance Feigl (1945, p. 285) suggests that this corresponds to a "'low-grade' explanation." Models of eye movement control, however, were "not intended to be a deep explanation of language processing, [...] because they do] not account for the many effects of higher-level linguistic processing on eye movements" (Reichle et al., 2003, p. 450).

Language models offer a deeper level of explanation, because they computationally specify how the prediction is generated. Therefore, they incorporate what can be called the three mnestic stages of the mental lexicon (cf. Paller and Wagner, 2002; Hofmann et al., 2018). All memory starts with experience, which is reflected by a text corpus (cf. Hofmann et al., 2020). The language models provide an algorithmic description of how long-term lexical knowledge is consolidated from this experience (Landauer and Dumais, 1997; Hofmann et al., 2018). Based on the consolidated syntactic and semantic lexical knowledge, language models are then exposed to the same materials that participants read and thus predict lexical retrieval. In the present study, we evaluate their predictions for viewing times during sentence reading (e.g., Staub, 2015).

We will compare CCP as a human-performance based explanation of reading against three types of language models. The probability that a word occurs, given two preceding words, is reflected in n-gram models, which capture syntactic and short-range semantic knowledge (cf. e.g., Kneser and Ney, 1995; McDonald and Shillcock, 2003a). This is a fully symbolic model, because the smallest unit of meaning representation consist of words. Second, we test topic models that are trained from word co-occurrence in documents, thus reflecting long-range semantics (Landauer and Dumais, 1997; Blei et al., 2003; Griffiths et al., 2007; Pynte et al., 2008a). Finally, recurrent neural networks (RNNs) most closely reflect the cloze completion procedure, because their hidden units are trained to predict a target word by all preceding words in a sentence (Elman, 1990; Frank, 2009; Mikolov, 2012). In contrast to the n-gram model, topic and RNN models distribute the meaning of a word across several subsymbolic units that do not represent human-understandable meaning by themselves.

## Eye Movements and Cloze Completion Probabilities

While the eyes sweep over a sequence of words during reading, they remain relatively still for some time, which is generally called fixation duration (e.g., Inhoff and Radach, 1998; Rayner, 1998). A very fast and efficient word recognition is obtained when a word can be recognized at a single glance. In this type of fixation event, the single-fixation duration (SFD) informs about this rapid and successful lexical access. When further fixations are required to recognize the word before the eyes move on to the next word, the duration of these further fixations is added to the gaze duration (GD). This is an eye movement measure that reflects "standard" lexical access in all words, while it may also represent syntactic and semantic integration (cf. e.g., Inhoff and Radach, 1998; Rayner, 1998; Radach and Kennedy, 2013). Finally, the eyes may come back to the respective word, and when the fixation times of these further fixations are added, this is reflected in the total viewing time (TVT)—an eye movement measure that reflects the full semantic integration of a word into the current language context (e.g., Radach and Kennedy, 2013).

Though CCP is known to affect all three types of fixation times, the result patterns considerably vary between studies (see e.g., Frisson et al., 2005; Staub, 2015; Brothers and Kuperberg, 2021, for reviews). A potential explanation is that CCP represents an all-in variable (Staub, 2015). The cloze can be completed because the word is expected from syntactic, semantic and/or event-based information—a term that refers to idiomatic expressions in very frequently co-occurring words (cf. Staub et al., 2015).

By shedding light on the consolidation mechanisms, language models are expected to complement future models of eye-movement control, which do not provide a deep explanation to linguistic processes (Reichle et al., 2003, p. 450). Models of eye-movement control, however, provide valuable insights how lexical access and eye-movements interact. These models assume that lexical access is primarily driven by word length, frequency and CCP-based predictability of the presently fixated word (e.g., Reichle et al., 2003; Engbert et al., 2005; Snell et al., 2018). This reflects the simplifying eye-mind assumption, which "posits that the interpretation of a word occurs while that word is being fixated, and that the eye remains fixated on that word until the processing has been completed" (Just and Carpenter, 1984, p. 169). Current models of eye-movement control, however, reject the idea that lexical processing exclusively occurs during the fixation of a word (Reichle et al., 2003; Engbert et al., 2005; see also Anderson et al., 2004; Kliegl et al., 2006). Lexical processing can still be underway when the eyes move on to the subsequent word, which can occur, for instance, if the first word is particularly difficult to process (Reilly and Radach, 2006). Therefore, lexical processing of the last word can still have a considerable impact on the viewing times of the currently fixated word. Moreover, when a word is currently fixated at a point in time, lexical access of the next word can already start (Reilly and Radach, 2006).

When trying to characterize the time course of single-word and contextual word properties, for instance the EZ-reader model suggests that there are two stages of lexical processing that are both influenced by word frequency and predictability. The first stage represents a "familiarity" check and the identification of an orthographic word form—this stage is primarily driven by word frequency. The second stage additionally involves the (phonological and) semantic word form—therefore, CCP has a stronger impact on this stage of processing. Please also note that attention can already shift to the next word, while the present word is fixated. When the next word is short, highly frequent and/or highly predictable, it can be skipped, and the saccade is programmed toward the word after the next word (Reichle et al., 2003).

## Language Models in Eye Movement Research

### Symbolic Representations in N-Gram Models

Symbolic n-gram models are so-called count-based models (Baroni et al., 2014; Mandera et al., 2017). Cases in which all n words co-occur are counted and related to the count of the preceding n-1 words in a text corpus. McDonald and Shillcock (2003a) were the first who tested whether a simple 2-gram model can predict eye movement data. They calculated the transitional probability that a word occurs at position n given the preceding word at position *n*-1. Then they paired preceding verbs with likely and less likely target nouns and showed significant effects on early SFD, but no effects on later GD (but see Frisson et al., 2005). Effects on GD were subsequently revealed using multiple regression analyses of eye movements, suggesting that 2-gram models also account for lexical access in all words (McDonald and Shillcock, 2003b; see also Demberg and Keller, 2008). McDonald and Shillcock (2003b) discussed that the 2-gram transitional probability reflects a relatively low-level process, while it does probably not capture high-level conceptual knowledge, corroborating the assumption that n-gram models reflect syntactic and short-range semantic information. Boston et al. (2008) analyzed the viewing times in the Potsdam Sentence Corpus (PSC, Kliegl et al., 2004) and found effects of transitional probability for all three fixation measures (SFD, GD, and TVT). Moreover, they found that these effects were descriptively larger than the CCP effects (see also Hofmann et al., 2017).

Smith and Levy (2013, p. 303) examined larger sequences of words by using a 3-gram model to show last- and present-word probability effects on GD during discourse reading (Kneser and Ney, 1995). Moreover, they showed that these n-gram probability effects are logarithmic (but cf. Brothers and Kuperberg, 2021). For their statistical analyses, Smith and Levy (2013) selected generalized additive models (GAMs) that can well capture the phenomenon that a predictor may perform better or worse in certain range of the predictor variable. They showed that the 3-gram probability of the last word still has a considerable impact on the GDs of the current word. Therefore, this type of language model can well predict that contextual integration of the last word is still underway at the fixation of the current word. Of some interest is that Smith and Levy (2013) suggest that CCP may predict reading performance well, when comparing extremely predictable with extremely unpredictable words. Hofmann et al. (2017, e.g., **Figure 3**), however, provide data showing that a 3-gram model may provide more accurate predictions at the lower end of the predictability distribution.

### Latent Semantic Dimensions

The best-known computational approach to semantics in psychology is probably latent semantic analysis (LSA, Landauer and Dumais, 1997). A factor-analytic-inspired approach is used to compute latent semantic dimensions that determine which words do frequently occur together in documents. This allows to address the long-range similarity of words and sentences by calculating the cosine distance (Deerwester et al., 1990). Wang et al. (2010) addressed the influence of transitional probability and LSA similarity of the target to the preceding content word. They found that transitional probability predicts lexical access, while the long-range semantics reflected by LSA particularly predicts late semantic integration [but see Pynte et al. (2008a,b) for LSA effects on SFD and GD, and Luke and Christianson (2016) for LSA effects on TVT, GD, and even earlier viewing time measures].

In a recent study, Bianchi et al. (2020) contrasted the GD predictions of an n-gram model with the predictions of an LSA-based match of the current word with the preceding nine words during discourse reading. They found that LSA did not provide effects over and above the n-gram model. The LSA-based predictions improved, however, when further adding the LSA-based contextual match of the next word. This indicates that such a document-level, long-range type of semantics might be particularly effective when taking the predictabilities of the non-fixated words into account.

LSA has been challenged by another dimension-reducing approach in accounting for eye movement data. Blei et al. (2003) introduced the topic model as a Bayesian, mere probabilistic language modeling alternative. Much as LSA, topic models are trained to reflect long-range relations based on the co-occurrence of words in documents. Griffiths et al. (2007) showed that topic models provide better model performance than LSA in many psychological tasks, such as synonym judgment or semantic priming. They calculated the probability of a word to occur, given topical matches with the preceding words in the sentence. This topic model-based predictor, but not LSA cosine accounted for Sereno's et al. (1992) finding that GDs and TVTs of a subordinate meaning are larger than in a frequency-matched non-ambiguous word (Sereno et al., 1992).

Though Hofmann et al. (2017) also found topic model effects on SFD data, their results suggested that long-range semantics provides comparably poor predictions. The short-range semantics and syntax provided by n-gram models, in contrast, provided a much better performance, particularly when the language models are trained by a corpus consisting of movie and film subtitles. In sum, the literature on document-level semantics presently provides no consistent picture. Long-range semantic effects might be comparably small (e.g., Hofmann et al., 2017), but they may be more likely to deliver consistent results when the analysis is not constrained to the long-range contextual match of the present, but also of other words (Bianchi et al., 2020). A more consistent picture might emerge, when also short-range predictability is considered, as reflected e.g., in n-gram models (Wang et al., 2010; Bianchi et al., 2020).

### (Recurrent) Neural Networks

Neural network models are deeply rooted in the tradition of connectionist modeling (e.g., Seidenberg and McClelland, 1989; McClelland and Rogers, 2003). In the last decade, these models were advanced in the machine learning community to successfully recognize pictures or machine translation (e.g., LeCun et al., 2015). In the processing of word stimuli, one of the most well-known of these models is the word2vec model, in which a set of hidden units is for instance trained to predict the surrounding words by the present word (Mikolov et al., 2013). This model is able to predict association ratings (Hofmann et al., 2018) or semantic priming (e.g., Mandera et al., 2017). The neural network that most closely approximates the cloze task, however, is the recurrent neural network model (RNN), because it is trained to predict the next word by the preceding sentence context. In RNN models, words are presented at an input layer, and a set of hidden units is trained to predict the probability of the next word at the output layer (Elman, 1990). The hidden layer is copied to a (recurrent) context layer after the presentation of each word. Thus, the network gains a computationally concrete form of short-term memory (Mikolov et al., 2013). Such a network provides large hidden-unit cosine distances between syntactic classes such as verbs and nouns, lower between non-living and living objects, and even lower between mammals and fishes, suggesting that RNNs reflect syntactic and short-range semantic information at the level of the sentence (Elman, 1990). Frank and Bod (2011) show that RNNs can account for syntactic effects in viewing times, because they absorb variance previously explainable by a hierarchical phrase-structure approach.

Frank (2009) used a simple RNN to successfully predict GDs during discourse reading. When adding transitional probability to their multiple regression analyses, both predictors revealed significant effects. Such a result demonstrates that prediction-based models such as RNNs and count-based n-gram models probably reflect different types of "predictability." Hofmann et al. (2017) showed that an n-gram model, a topic model, and an RNN model together can significantly outperform CCP for the prediction of SFD. It is, however, unclear whether this finding can be replicated in a different data set and generalized to other viewing time measures.

Some recent studies compared other types of neural network models to CCP (Bianchi et al., 2020; Wilcox et al., 2020; Lopukhina et al., 2021). For example, Bianchi et al. (2020) explored the usefulness of word2vec. Because they did not find stable word2vec predictions for eye movement data, they decided against a closer examination of this approach. Rather they relied on fasttext—another non-recurrent neural model, in which the hidden units are trained to predict the present word by the surrounding language context (Mikolov et al., 2018). Moreover, Bianchi et al. (2020) evaluated the performance of an n-gram model and LSA. When comparing the performance of these language models, they obtained the most reliable GD predictions for their n-gram model, followed by CCP, while LSA and fasttext provided relatively poor predictions. In sum, studies comparing CCP to language models support the view that CCP-based and language-model-based predictors account for different though partially overlapping variances in eye-movement data (Bianchi et al., 2020; Lopukhina et al., 2021) that seem related to syntactic, as well as early and late semantic processing during reading.

### The Present Study

The present study was designed to overcome the limitations of the pilot study of Hofmann et al. (2017), which compared an n-gram, a topics and an RNN model with respect to the prediction of CCP, electrophysiological and SFD data in only the PSC data set. They found that RNN models and n-gram models provide a similar performance in predicting these data, while the topics model made remarkably worse predictions. In the present study, we focused on eye movements and aimed to replicate the SFD effects with a second sample, which was published by Schilling et al. (1998). Moreover, we aimed to examine the dynamics of lexical processing. By modeling a set of three viewing time parameters (SFD, GD and TVT), we will be able to compare the predictions of CCP and different language models for early rapid (SFD) and standard (GD) lexical access, and their predictions for full semantic integration (TVT). In their linear multiple regression analysis on item-level data, Hofmann et al. (2017) found that the three language models together account for around 30% of reproducible variance in SFD data—as opposed to 18% for the CCP model. Though the three language models together significantly outperformed the CCP-based approach, they used Fisher-Yates significance $z$-to-$t$-tests as a conservative approach, because aggregating over items results in a strong loss of variance. Therefore, n-gram and RNN models alone outperformed CCP always at a descriptive level, but the differences were not significant. Here we applied a model comparison approach to evaluate the model fit in comparison to a baseline model, using standard log likelihood tests (e.g., Baayen et al., 2008). This approach will also test the assumptions of different short- and long-range syntactic and semantic processes that we expect to be reflected by the parameters of the three different language models selected.

Such a statistical approach, however, is based on unaggregated data. As Lopukhina et al. (2021) pointed out, the predictability effects in such analyses are relatively small. For instance Kliegl et al. (2006; cf. **Table 5**) found that CCP can account for up to 0.38% of the variance in viewing times – thus it is important to evaluate the usefulness of language models in highly powered samples. On the other hand, a smaller sample reflects a more typical experimental situation. The present study was designed to replicate and extend previous analyses of viewing time parameters using two independent eye-movement data sets, a very large sample of CCP and eye movement data, the PSC, and a sample that is more typical for eye movement experiments, the SRC.

In addition to a simple item-level analysis as a standard benchmark for visual word recognition models (Spieler and Balota, 1997), that were applied more thoroughly in a previous set of analyses (Hofmann et al., 2017), we here applied Smith and Levy's (2013) generalized additive modeling approach with a logarithmic link function (but cf. Brothers and Kuperberg, 2021). The computed GAMs rely on fixation-event-level viewing time parameters as the dependent variables. We used a standard set of baseline predictors for reading and lexical access, and then extended this baseline model by CCP- and/or language-model-based predictors for the present, last and next words. To test for reproducibility, our analyses will be based on the two eye-movement data sets that are most frequently used for testing models of eye-movement control: the EZ-reader model (Reichle et al., 2003) was tested with the SRC data set; and the SWIFT and the OB-1 reader models were used to predict viewing times in the PSC (Kliegl et al., 2004; cf. Engbert et al., 2005; Snell et al., 2018). GAMs are non-linear extensions of the generalized linear models that allow predictors to be modeled as a sum of smooth functions and therefore allow better adaptations to curvilinear and wiggly predictor-criterion relationships (Wood, 2017).

Different language models are expected to explain differential and independent proportions of variance in the viewing time parameters. While an n-gram model reflects short-range semantics, we expect it to be predictor of all viewing time measures (e.g., Boston et al., 2008). A subsymbolic topic model that reflects long-range semantics should be preferred over the other language models in predicting GD and TVT and semantic integration into memory (Sereno et al., 1992; Griffiths et al., 2007), particularly when other forms of predictability are additionally taken into account (Wang et al., 2010; Bianchi et al., 2020). Previous studies examining RNN models found effects on SFD and GD (e.g., Frank, 2009; Hofmann et al., 2017). Thus, it is an open empirical question whether predict-based models do not only affect lexical access, but also late semantic integration. As these models are trained to predict the next word, they may be particularly useful to examine early lexical preprocessing of the next word.

## METHOD
### Language Model Simulations

All language models were trained by **corpora** derived from movie and film subtitles.[1] The English Subtitles training corpus consisted of 110 thousand films and movies that were used for document-level training of the topic models. We used the 128 million utterances as the sentence-level, in order to train the n-gram and RNN models in the English corpus, which in all consisted of 716 million tokens. The German corpus consisted of 7 thousand movies, 7 million utterances/sentences comprising of 54 million tokens.

Statistical **n-gram** models for words are defined by a sequence of $n$ words, in which the probability of the $n^{th}$ word depends on a Markov chain of the previous $n$-1 words (see, e.g., Chen and Goodman, 1999; Manning and Schütze, 1999). Here we set $n = 3$ and thus computed the conditional probability of a word $w_n$, given the two previous words ($w_{n-1} \ldots w_1$; Smith and Levy, 2013).

$$p(w_n | w_1 \ldots w_{n-1}) = \frac{p(w_1 \ldots w_n)}{p(w_1 \ldots w_{n-1})} \quad (1)$$

We used Kneser-Ney-smoothed 3-gram models, relying on the BerkeleyLM implementation (Pauls and Klein, 2011).[2] These models were trained by the subtitles corpora to capture lexical memory consolidation (cf. Hofmann et al., 2018). For modeling lexical retrieval, we computed the conditional probabilities for the sentences presented in the SRC and the PSC data set (cf. below). Since n-gram models only rely on the most recent history for predicting the next word, they fail to account for longer-range phenomena and semantic coherence (see Biemann et al., 2012).

For training the **topic** models, we used the procedure by Griffiths and Steyvers (2004), who infer per-topic word distributions and per-document topic distributions through a Gibbs sampling process. The empirically observable probability of a word $w$ to occur in a document $d$ is thus approximated by the sum of the products of the probabilities a word, given the respective topic $z$, and the topic, given the respective word:

$$p = \sum_{i=1 \ldots N} p(w | z_i) * p(z_i | d) \quad (2)$$

Therefore, words frequently co-occurring in the same documents receive a high probability in the same topic. We use Phan and Nguyen's (2007) Gibbs-LDA implementation[3] for training a latent dirichlet allocation (LDA) model with $N = 200$ topics (default values for $\alpha = 0.25$ and $\beta = 0.001$; Blei et al., 2003). The per-document topic distributions are trained in form of a topic-document matrix [$p(z_i|d)$], allowing to classify documents by topical similarities, and used for inference of new (unseen) "documents" at retrieval.

For modeling lexical retrieval of the SRC and PSC text samples, we successively iterate over the words of the particular sentence and create a new LDA document representation $d$ for each word at time $i$ and its entire history of words in the same sentence:

$$p(w_i | d) = p(w_i | w_i \ldots w_1) \quad (3)$$

In this case, $d$ refers to the history of the current sentence including the current word $w_i$, where we are only interested in the probability of $w_i$. We here computed the probabilities of the current word $w_i$ given its history as a mixture of its topical components (cf. Griffiths et al., 2007, p. 231f), and thus address the topical matches of the present word with the preceding words in the sentence context.

---
[1] www.opensubtitles.org

[2] https://code.google.com/p/berkeleylm/
[3] http://gibbslda.sourceforge.net/

For the **RNN** model simulations, we relied on the faster RNN-LM implementation[4], which can be trained on huge data sets and very large vocabularies (cf. Mikolov, 2012). The input and target output units consist of so-called one-hot vectors with one entry for each word in the lexicon of this model. If the respective word is present, the entry corresponds to 1, while the entries remain 0 for all other words. At the input level, the entire sentence history is given word-by-word and the models objective is to predict the probability of the next word at the output level. Therefore, the connection weights of the input and output layer to the hidden layer are optimized. At model initialization, all weights are assigned randomly. As soon as the first word is presented to the input layer, the output probability of the respective word unit is compared to the actual word, and the larger the difference, the larger will be the connection weight change (i.e., backpropagation by a simple delta rule). When the second word of a sentence then serves as input, the state of the hidden layer after the first word is copied to a context layer (cf. **Figure 2** in Elman, 1990). This (recurrent) context layer is used to inform the current prediction. Therefore, the RNN obtains a form of short-term memory (Mikolov, 2012; cf. Mikolov et al., 2013). We trained a model with 400 hidden units and used the hierarchical softmax provided by faster-RNN with a temperature of 0.6, using a sigmoid activation function for all layers. For computing lexical retrieval, we used the entire history of a sentence up to the current word and computed the probability for that particular word.

## Cloze Completion and Eye Movement Data

The CCP and eye movement data of the SRC and the PSC were retrieved from Engelmann et al. (2013).[5] The SRC data set contains incremental cloze task and eye movement data for 48 sentences and 536 words that were initially published by Schilling et al. (1998). The PSC data set provides the same data for 144 sentences and 1,138 words (Kliegl et al., 2004, 2006).

The sentence length of the PSC ranges from 5 to 11 words ($M = 7.9$; $SD = 1.39$) and from 8 to 14 words in the SRC ($M = 11.17$; $SD = 1.36$). As last-word probability cannot be computed for the first word in a sentence, and next-word probability cannot be computed for the last word of a sentence, we excluded fixation durations on the first and the last words of each sentence from analyses. Four words of the PSC (e.g., "Andendörfern," villages of the Andes) did not occur in the training corpus and were excluded from analyses. This resulted in the 440 target words for the SRC and the 846 target words for the PSC analyses. The respective participant sample sizes and number of sentences are summarized in **Table 1** (see Schilling et al., 1998; Kliegl et al., 2004, 2006, for further details). **Table 2** shows example sentences, in which one type of predictability is higher than the other predictability scores. In general, CCP distributes the probability space across a much smaller number of potential completion candidates. Therefore, the mean probabilities are comparably high (SRC: $p = 0.3$; PSC: $p = 0.2$). The mean of the computed predictability scores, in contrast, provide 2–3 leading zeros. Moreover, the computed predictability scores by far provide greater probability ranges.

To compute SRC-based *CCP* scores comparable to the PSC (Kliegl et al., 2006), we used the empirical cloze completion probabilities (*ccp*) and logit-transformed them (*CCP* in formula 4). Because Kliegl's et al. (2004) sample was based on 83 complete predictability protocols, cloze completion probabilities of 0 and 1 were replaced by 1/(2*83) and 1−[1/(2*83)] for the SRC, to obtain the same extreme values.

$$CCP = 0.5 * \log\left(\frac{ccp}{1 - ccp}\right) \qquad (4)$$

Since lexical processing efficiency varies with landing position of the eye within a word (e.g., O'Regan and Jacobs, 1992; Vitu et al., 2001), we computed relative landing positions by dividing the landing letter by the word length. The optimal viewing position is usually slightly left to the middle of the word, granting optimal visual processing of the word (e.g., Nuthmann et al., 2005). Therefore, we will use the landing position as a covariate to partial out variance explainable by suboptimal landing positions (cf. e.g., Vitu et al., 2001; Kliegl et al., 2006; Pynte et al., 2008b). For all eye movement measures, we excluded fixation durations below 70 ms (e.g., Radach et al., 2013). The upper cutoff was defined by examining the data distributions and excluding the range in which only a few trials remained for analyses. We excluded durations 800 ms or greater for SFD (21 fixation durations for SRC and 13 for PSC), 1,200 ms for GD (12 for SRC and 0 for PSC), and 1,600 ms for TVT analyses (7 for SRC and 0 for PSC). This resulted in the row numbers used for the respective analyses given in **Table 1**.

## Data Analysis

First, we calculated simple linear item-level correlations between the predictor variables and the mean SFD, GD and TVT data (see **Table 3**). In addition to the logit-transformed CCPs and the $\log_{10}$-transformed language model probabilities (Kliegl et al., 2006; Smith and Levy, 2013), we also explored the correlations of the non-transformed probability values with SFD, GD and TVT data, respectively: In the SRC data set, CCP provided correlations of −0.28, −0.33, and −0.39; n-gram models of −0.11, −0.16 and −0.21; topic models of −0.35, −0.47 and −0.52; and RNN models provided correlations of −0.16, −0.23, and −0.25, respectively. In the PSC data set, the SFD, GD and TVT correlations with CCP were −0.20, −0.26, −0.31; those of n-gram models were −0.16, −0.18 and −0.19; topics models provided correlations of −0.19, −0.18 and −0.17; and RNN models of −0.19, −0.21, −0.22. In sum, the transformed probabilities always provided higher correlations with all fixation durations than the untransformed probabilities (cf. **Table 3**). Therefore, the present analyses focus on transformed values.

For non-linear fixation-event based analyses of the non-aggregated eye-movement data, we relied on GAMs using thin

---

[4] https://github.com/yandex/faster-rnnlm
[5] https://clarinoai.informatik.uni-leipzig.de/fedora/objects/mrr:11022000000001F2FB/datastreams/EngelmannVasishthEngbertKliegl2013_1.0/content

**TABLE 1** | Overview about cloze completion and eye movement (EM) data used for the present study.

| Data set | Sentences | Targets | Language | Participants | | Rows of data in analysis | | |
|---|---|---|---|---|---|---|---|---|
| | | | | CCP | EM | SFD | GD | TVT |
| SRC | 48 | 440 | English | 20 | 30 | 6,451 | 8,671 | 8,736 |
| PSC | 144 | 846 | German | 272 | 222 | 100,975 | 134,835 | 135,021 |

**TABLE 2** | Example sentences and the probabilities of the four types of predictability.

| SRC | | | | | PSC | | | | |
|---|---|---|---|---|---|---|---|---|---|
| Word | CCP | N-gram | Topic | RNN | Word | CCP | N-gram | Topic | RNN |
| Bill | 6e-3 | 1e-4 | 1e-3 | 2e-5 | In | 1e-2 | 2e-3 | 3e-2 | 4e-3 |
| complained | 6e-3 | 3e-6 | 1e-4 | 1e-6 | der | 7e-1 | 1e-1 | 1e-2 | 1e-1 |
| that | 3e-1 | 1e-1 | 2e-3 | 6e-2 | Klosterschule | 6e-3 | 2e-6 | 4e-5 | 4e-5 |
| the | 2e-1 | 1e-1 | 2e-3 | 1e-2 | herrschen | 2e-2 | 1e-6 | 8e-4 | 3e-4 |
| magazine | 6e-3 | 1e-4 | 3e-4 | 3e-5 | **Schwester** | 6e-3 | 5e-5 | **4e-2** | 1e-9 |
| included | 6e-3 | 3e-7 | 2e-4 | 1e-5 | Agathe | 1e-2 | 7e-7 | 1e-4 | 1e-8 |
| more | 6e-3 | 4e-4 | 2e-3 | 6e-4 | und | 9e-1 | 5e-3 | 4e-3 | 4e-2 |
| adds | **4e-1** | 6e-7 | 2e-5 | 1e-8 | Schwester | 5e-1 | 1e-4 | 2e-2 | 8e-5 |
| than | 9e-1 | 2e-4 | 2e-3 | 1e-3 | Maria | 1e-1 | 5e-4 | 2e-3 | 2e-3 |
| articles | 8e-1 | 4e-6 | 5e-5 | 4e-6 | | | | | |
| The | 6e-3 | 6e-4 | 3e-3 | 2e-2 | Er | 6e-3 | 1e-2 | 2e-2 | 2e-2 |
| drunk | 6e-3 | 6e-5 | 9e-4 | 2e-5 | hätte | 6e-3 | 5e-3 | 2e-3 | 2e-3 |
| **driver** | 6e-3 | **2e-2** | 2e-4 | 2e-3 | nicht | 2e-2 | 4e-2 | 2e-2 | 3e-2 |
| lost | 6e-3 | 6e-6 | 2e-3 | 1e-5 | auch | 6e-3 | 3e-4 | 9e-4 | 4e-3 |
| control | 4e-1 | 4e-1 | 3e-3 | 6e-3 | noch | 7e-1 | 1e-1 | 1e-3 | 3e-2 |
| crashed | 5e-2 | 9e-8 | 1e-4 | 4e-7 | am | 1e-2 | 3e-3 | 3e-3 | 5e-3 |
| into | 4e-1 | 9e-2 | 2e-3 | 1e-1 | **Telefon** | 6e-3 | 4e-3 | 4e-3 | **4e-2** |
| a | 6e-1 | 2e-1 | 2e-3 | 1e-1 | nörgeln | 6e-3 | 8e-8 | 5e-5 | 6e-8 |
| street | 6e-3 | 9e-4 | 2e-3 | 4e-3 | sollen | 7e-1 | 2e-4 | 2e-3 | 6e-4 |
| sign | 6e-1 | 2e-2 | 3e-3 | 6e-4 | | | | | |
| and | 8e-1 | 1e-3 | 2e-3 | 3e-3 | | | | | |
| died | 7e-1 | 3e-5 | 2e-3 | 4e-4 | | | | | |
| *M* | 3e-1 | 5e-2 | 1e-3 | 3e-2 | *M* | 2e-1 | 2e-2 | 8e-3 | 1e-2 |
| *SD* | 4e-1 | 1e-1 | 1e-3 | 7e-2 | *SD* | 3e-1 | 7e-2 | 2e-2 | 3e-2 |
| *Min* | 6e-3 | 1e-9 | 3e-6 | 2e-10 | *Min* | 6e-3 | 1e-10 | 2e-6 | 4e-13 |
| *Max* | 1e+0 | 1e+0 | 2e-2 | 5e-1 | *Max* | 1e+0 | 9e-1 | 2e-1 | 5e-1 |

*Examples sentences were selected to illustrate one case, in which one type of predictability is particularly high (bold). Translations (PSC): In the convent school, nun Agathe, and nun Maria rule (upper). He should not have moaned at the telephone, as well (lower sentence).*

plate regression splines from the mgcv-package (version 1.8) in R (Hastie and Tibshirani, 1990; Wood, 2017). As several models of eye-movement control rely on the gamma distribution (Reichle et al., 2003; Engbert et al., 2005), we here also used gamma functions with a logarithmic link function (cf. Smith and Levy, 2013). GAMs have the advantage to model non-linear smooth functions, i.e., the GAM aims to find the best value for the smoothing parameter in an iterative process. Because smooth functions are modeled by additional parameters, the amount of smoothness is penalized in GAMs, i.e., the model aims to reduce the number of parameters of the smooth function and thus to avoid overfitting. The effective degrees of freedom (edf) parameter describes the resulting amount of smoothness (see **Table 8** below). Of note is, that an edf of 1 is present if the model penalized the smooth term to a linear relationship. Edf's close to 0 indicate that the predictor has zero wiggliness and can be interpreted to be penalized out of the model (Wood, 2017). Though Baayen (2010) suggested that word frequency can be seen as a collector variable that actually also contains variance from contextual word features (cf. Ong and Kliegl, 2008), our baseline GAMs contained single-word properties. We computed a baseline GAM consisting of the length and frequency of the present, last and next word as predictors (cf. Kliegl et al., 2006). To reduce the correlations between the language models trained by the subtitles corpora and the frequency measures, word frequency estimates were taken from

TABLE 3 | Correlations between word properties for the SRC (below diagonal) and PSC (above diagonal), and the item-level means of the SFD, GD, and TVT data of the present word.

|  | 1 | 2. | 3 | 4 | 5 | 6 | 7 | 8 | 9 |
|---|---|---|---|---|---|---|---|---|---|
| 1. Length |  | −0.62 | −0.40 | −0.47 | −0.46 | −0.51 | 0.28 | 0.62 | 0.57 |
| 2. Frequency | −0.76 |  | 0.52 | 0.71 | 0.70 | 0.75 | −0.35 | −0.49 | −0.50 |
| 3. CCP | −0.48 | 0.58 |  | 0.56 | 0.36 | 0.56 | **−0.27** | **−0.34** | **−0.40** |
| 4. N–gram | −0.58 | 0.74 | 0.63 |  | 0.61 | 0.79 | *−0.41* | *−0.49* | *−0.51* |
| 5. Topic | −0.67 | 0.80 | 0.45 | 0.69 |  | 0.61 | *−0.35* | *−0.43* | *−0.41* |
| 6. RNN | −0.65 | 0.81 | 0.58 | 0.84 | 0.75 |  | *−0.47* | *−0.51* | *−0.53* |
| 7. SFD | 0.35 | −0.50 | **−0.33** | *−0.39* | *−0.40* | *−0.44* |  | 0.81 | 0.79 |
| 8. GD | 0.54 | −0.62 | **−0.38** | *−0.51* | *−0.55* | *−0.55* | 0.86 |  | 0.95 |
| 9. TVT | 0.61 | −0.66 | **−0.44** | *−0.55* | *−0.58* | *−0.60* | 0.78 | 0.90 |  |

*Highlighting was used to illustrate that the language models (italics and bold) provide always larger correlations with the three viewing time measures than CCP (bold) (see below for discussions).*

TABLE 4 | Generalized additive models (GCV, $R^2$) for single-fixation duration (SFD) and $\chi^2$ tests (df) against the previous model for significant increments in explained variance ($^*p < 0.05$).

|  | SRC | | | PSC | | |
|---|---|---|---|---|---|---|
|  | GCV | %Δ$R^2$ | Deviance (df) | GCV | %Δ$R^2$ | Deviance (df) |
| Baseline | 0.1273 | 4.17 | Baseline | 0.0969 | 3.99 | Baseline |
| **CCP** | **Baseline** |  |  |  |  |  |
| Baseline + present | 0.1273 | 0.02 | 0.4 (1.9) | 0.0968 | 0.1 | 10.7 (8)* |
| + Last | **0.1272** | **0.06** | **1.1 (2.7)*** | 0.0967 | 0.05 | 7.5 (9.1)* |
| + Next | 0.1271 | 0.02 | 0.5 (1.1) | 0.0964 | 0.31 | 36.4 (9.5)* |
| **N-gram** | **Baseline** |  |  |  |  |  |
| Baseline + present | **0.127** | **0.29** | **3.1 (5.6)*** | 0.0966 | 0.27 | 33 (9.1)* |
| + Last | **0.1265** | **0.44** | **5.5 (9.4)*** | 0.0964 | 0.14 | 15.5 (8.7)* |
| + Next | 0.1265 | −0.02 | 0 (0.8) | 0.0963 | 0.07 | 9.6 (8.5)* |
| **Topic** | **Baseline** |  |  |  |  |  |
| Baseline + present | 0.1272 | 0.12 | 1.4 (4.6) | 0.0967 | 0.21 | 23.1 (8.4)* |
| + Last | 0.1272 | 0.04 | 1.1 (5.8) | 0.0963 | 0.28 | 33.8 (9.2)* |
| + Next | 0.1272 | 0.09 | 1.3 (4.4) | 0.0963 | 0.07 | 9.6 (9.2)* |
| **RNN** | **Baseline** |  |  |  |  |  |
| Baseline + present | **0.1271** | **0.16** | **1.6 (1.2)*** | 0.0966 | 0.27 | 31.3 (9.6)* |
| + Last | 0.1271 | −0.02 | 0 (0.9) | 0.0966 | 0.03 | 3.7 (8.7)* |
| + Next | **0.1269** | **0.27** | **3.6 (11)*** | 0.0964 | 0.16 | 18 (8.7)* |
| **N-gram + Topic + RNN** | **Full CCP model** |  |  |  |  |  |
| (Present + last + next) | **0.1262** | **1.08** | **13 (33.6)*** | 0.0957 | 0.69 | 81.5 (49.1)* |

*Consistent GAM model improvements in both data sets are marked bold.*

the Leipzig corpora collection[6] The English corpus consisted of 105 million unique sentences and 1.8 billion words, and the German corpus consisted of 70 million unique sentences and 1.1 billion words (Goldhahn et al., 2012). We used Leipzig word frequency classes that relate the frequency of each word to the frequency of the most frequent word using the definition that the most common word is $2^{class}$ more frequent than the word of which the frequency is given ("der" in German and "the" in English; e.g., Hofmann et al., 2011, 2018). Moreover, we inserted landing site into the baseline GAM (e.g., Pynte et al., 2008b), to absorb variance resulting from mislocated fixations.

We added the different types of predictability of the present word to the baseline model and tested whether the resulting GAM performs better than the baseline GAM (**Tables 4–6**). Then we successively added the predictability scores of the last and next words and tested whether the novel GAM performs better than the preceding GAM. Finally, we also tested whether a GAM model including all language-model-based predictors provides better predictions than the GAM including CCP scores (Hofmann et al., 2017).

---

[6] http://www.corpora.uni-leipzig.de/en?corpusId=deu_newscrawl-public_201

**TABLE 5** | Generalized additive models (GCV, $R^2$) for gaze duration (GD) and $\chi^2$ tests (df) against the previous model for significant increments in explained variance (*$p <$ 0.05).

|  | SRC | | | PSC | | |
|---|---|---|---|---|---|---|
|  | GCV | %Δ$R^2$ | Deviance (df) | GCV | %Δ$R^2$ | Deviance (df) |
| Baseline | 0.1656 | 8.19 | Baseline | 0.145 | 10.52 | Baseline |
| **CCP** | **Baseline** | | | | | |
| Baseline + present | 0.1656 | 0.02 | 0.5 (1.6) | 0.1448 | 0.09 | 31.2 (8.3)* |
| + Last | **0.1654** | **0.14** | **5 (11.1)*** | 0.1447 | 0.05 | **14.2 (9)*** |
| + Next | **0.1652** | **0.08** | **2 (0.6)*** | 0.1444 | 0.17 | **45.8 (9.4)*** |
| **N-gram** | **Baseline** | | | | | |
| Baseline + present | **0.1651** | **0.32** | **4.3 (0.9)*** | 0.1446 | 0.24 | **66.8 (9)*** |
| + Last | **0.1648** | **0.14** | **3.8 (5.2)*** | 0.1443 | 0.13 | **34.1 (8.8)*** |
| + Next | 0.1648 | 0.02 | 1.1 (5.7) | 0.1443 | 0.04 | 13.5 (8.7)* |
| **Topic** | **Baseline** | | | | | |
| Baseline + present | 0.1657 | 0.01 | −0.6 (0.6) | 0.1448 | 0.15 | 38.1 (7.7)* |
| + Last | 0.1656 | 0.06 | 2.4 (6.9) | 0.1444 | 0.14 | 47 (9.1)* |
| + Next | 0.1656 | 0 | 0.1 (1.2) | 0.1444 | 0.02 | 10.2 (9)* |
| **RNN** | **Baseline** | | | | | |
| Baseline + present | **0.1653** | **0.21** | **4 (5.3)*** | 0.1446 | 0.22 | **64.1 (8.9)*** |
| + Last | 0.1652 | 0.1 | 2.2 (5.7) | 0.1446 | 0.02 | 6.6 (7.9)* |
| + Next | **0.1651** | **0.09** | **2.4 (4.8)*** | 0.1444 | 0.12 | **26 (7.8)*** |
| **N-gram + Topic + RNN** | **Full CCP model** | | | | | |
| (Present + last + next) | **0.1644** | **0.67** | **14.1 (28.2)*** | 0.1434 | 0.55 | **145.8 (52)*** |

*Consistent GAM model improvements in both data sets are marked bold.*

**TABLE 6** | Generalized additive models (GCV, $R^2$) for total viewing time (TVT) and $\chi^2$ tests (df) against the previous model for significant increments in explained variance (*$p <$ 0.05).

|  | SRC | | | PSC | | |
|---|---|---|---|---|---|---|
|  | GCV | %Δ$R^2$ | Deviance (df) | GCV | %Δ$R^2$ | Deviance (df) |
| Baseline | 0.1933 | 9.73 | Baseline | 0.1952 | 9.94 | Baseline |
| **CCP** | **Baseline** | | | | | |
| Baseline + present | **0.1931** | **0.13** | **2.8 (2.2)*** | 0.1943 | 0.32 | **134.4 (8.3)*** |
| + Last | 0.1931 | 0.02 | 0.6 (1.9) | 0.194 | 0.14 | 42.8 (9.3)* |
| + Next | **0.1929** | **0.14** | **4.2 (7.8)*** | 0.1938 | 0.09 | **32.6 (9.2)*** |
| **N-gram** | **Baseline** | | | | | |
| Baseline + present | **0.1926** | **0.39** | **8.8 (7.2)*** | 0.1942 | 0.33 | **139.6 (8.9)*** |
| + Last | **0.1925** | **0.09** | **2.7 (5.5)*** | 0.1937 | 0.23 | **81.8 (8.9)*** |
| + Next | 0.1925 | 0.01 | 1.7 (5.6) | 0.1935 | 0.07 | 26.4 (8.8)* |
| **Topic** | **Baseline** | | | | | |
| Baseline + present | **0.1932** | **0.15** | **4.1 (7.9)*** | 0.1949 | 0.12 | **48.2 (8.8)*** |
| + Last | 0.1932 | 0 | 2.1 (8.1) | 0.1945 | 0.15 | 56.8 (8.8)* |
| + Next | 0.1933 | −0.01 | 0 (0.9) | 0.1944 | 0.03 | 17.1 (9.3)* |
| **RNN** | **Baseline** | | | | | |
| Baseline + present | **0.1926** | **0.33** | **6.1 (0.2)*** | 0.1942 | 0.34 | **138.2 (8.2)*** |
| + Last | **0.1925** | **0.12** | **3.4 (6.9)*** | 0.194 | 0.1 | **32.5 (8.7)*** |
| + Next | **0.1923** | **0.16** | **5.4 (11.5)*** | 0.1939 | 0.04 | **18.8 (8.8)*** |
| **N-gram + Topic + RNN** | **Full CCP model** | | | | | |
| (Present + last + next) | **0.1917** | **0.64** | **16.6 (19.9)*** | 0.1924 | 0.52 | **206 (53.5)*** |

*Consistent GAM model improvements in both data sets are marked bold.*

TABLE 7 | $\chi^2$ tests whether the respective language model performed better than CCP.

| | SRC | | | | | | PSC | | | | | |
|---|---|---|---|---|---|---|---|---|---|---|---|---|
| | SFD | | GD | | TVT | | SFD | | GD | | TVT | |
| | %$\Delta R^2$ | Deviance (df) | %$\Delta R^2$ | Deviance (df) | %$\Delta R^2$ | Deviance (df) | %$\Delta R^2$ | Deviance (df) | %$\Delta R^2$ | Deviance (df) | %$\Delta R^2$ | Deviance (df) |
| N-gram | **0.6** | **6.5 (10.1)*** | 0.24 | 1.8 (−1.4) | **0.21** | **5.6 (6.5)*** | 0.02 | 3.6 (−0.2) | 0.09 | 23.2 (−0.2) | 0.09 | 38 (−0.2) |
| Topic | 0.14 | 1.8 (9) | −0.17 | −5.7 (−4.6)* | −0.15 | −1.4 (5) | **0.1** | **12 (0.2)*** | 0 | 4 (−1) | −0.24 | −87.7 (0) |
| RNN | **0.31** | **3.2 (7.4)*** | 0.17 | 1 (2.5) | **0.33** | **7.3 (6.7)*** | 0 | −1.6 (0.3) | 0.04 | 5.4 (−2.1) | *−0.06* | *−20.3 (−1.1)** |

*Positive deviance (df) suggests better performance of the language model, and negative deviance indicates that CCP fits better (*p < 0.05).*
*For a better overview, language models performing better were marked bold, and CCP performing better was marked in italics and bold.*

As model benchmarks, we report the generalized cross-validation score (GCV). This is an estimate of the mean prediction error based on a leave-one-out cross validation process. Better models provide a lower GCV (Wood, 2017). We also report the difference in the percentage of explained variance relative to the preceding or baseline model (%$\Delta R^2$, derived from adjusted $R^2$-values). We also tested whether a subsequent GAM provides significantly greater log likelihood than the previous model using $\chi^2$-tests (anova function in R; $p = 0.05$, cf. **Tables 4–6**). To provide a measure that can be interpreted in a similar fashion as the residual sum of squares for linear models, we further report the difference of the deviance of the last and the present model (e.g., Wood, 2017). If this term is negative, this indicates that the latter model provides a better account for the data. We also report the difference of the degrees of freedom (df) of the models to be compared. Negative values indicate that the previous GAM is more complex.

In the second set of GAM comparisons, we compare the performance of each single language model to the performance of the CCP. For this purpose, we use the predictability scores for all positions (present, last, and next word), and compared each language model to CCP (see **Table 7** below). To examine the predictors themselves and to be able to directly compare the contribution of human-generated and model-based predictors in explaining variance in viewing times, we also generated a final GAM model for each viewing time parameter comprising all types of predictability. For these models we finally report the $F$-values, effective degrees of freedom and the levels of significance (cf. **Table 8**). We evaluate the functional forms of the effects that are most reproducible across all analyses in the final model, while setting all non-inspected variables to their mean value (cf. **Figures 1–3** below).

## RESULTS

Our **simple item-level correlations** revealed that all language models provided larger correlations with SFD, GD, and TVT data than CCP (**Table 3**), demonstrating that language models provide a better account for viewing times than CCP. Moreover, there are substantial correlations between all predictor variables, making analyses with multiple predictors prone to fit error variance. Therefore, we will focus our conclusions on those findings that can be reproduced in different types of analyses (**Tables 4–8**; Wagenmakers et al., 2011). When turning to these **non-linear GAM analyses** at the level of each fixation event, we found that nearly any predictor accounts for variance in the PSC data set. This suggests that all types of predictability account for viewing time variance, once there is sufficient statistical power. When we examined typically sized samples in the SRC, only the most robust effects make a contribution. Therefore, we will also focus our conclusions on those effects that can be reproduced in both samples (see **Table 8** for a summary of all results).

Concerning the **CCP** analyses, the only findings that can be reproduced in both data samples and across all non-linear analyses was the influence of last- and next-word CCP on GD data (**Tables 5, 8**). These effects seem quite robust and can be examined in **Figure 1**. When all types of predictability are included in the GAM (**Table 8**), CCP of the last and next word seems to prolong GDs particularly in the range of logit-transformed CCPs of around 0.5–1.5 (**Figure 1**). We see this as a preliminary evidence that this type of predictability might be particularly prone to predict that high-CCP last or next words are processed during present-word fixations.

In general, CCP effects seem to be most reliable in GD data. CCP outperformed the topic model in the SRC data set for GD predictions (**Table 7**). Please also note that the only type of predictability that failed to improve the GAM predictions in the highly powered PSC sample was the CCP-effect of the last word in SFD data (**Table 8**).

The language models not only showed greater correlations with all viewing time measures than CCP (**Table 3**), but they also delivered a larger number of consistent findings in our extensive set of non-linear analyses (**Tables 4–8**). There are some CCP effects worth to be highlighted that are exclusively apparent in the analyses using only a single type of predictability. There are last-word CCP effects in SFD data (**Table 4**), and a present-word effect in the TVT data (**Table 6**). In addition to the fully consistent last-word GD effects (**Tables 5, 8, Figure 1**), this lends further support to the hypothesis that CCP is particularly prone to reflect late processes. Moreover, there are also consistent next-word effects in the GD and TVT analyses of both samples (**Tables 5, 6**) that are, however, often better explained by other types of predictability in the analysis containing all predictors (**Table 8**).

Our non-linear analyses revealed that the **n-gram**-based probabilities of the present word can account for variance in all

TABLE 8 | The F-values of the predictors (effective degrees of freedom), their significance level and the total amount of variance explained in models including all predictors at the same time.

| Baseline | SFD | | GD | | TVT | |
|---|---|---|---|---|---|---|
| | SRC | PSC | SRC | PSC | SRC | PSC |
| Landing site | 2.9 (1.9)* | 12.8 (8.6)*** | 8.4 (4.1)*** | 12.8 (8.7)*** | 10.5 (4.5)*** | 10.7 (8.5)*** |
| Length | 0.6 (2.0) | 19.6 (8.8)*** | 10.2 (3.2)*** | 372.5 (9.0)*** | 17.2 (2.2)*** | 248.9 (9.0)*** |
| Length_last | 6.9 (7.0)*** | 29.1 (8.9)*** | 6.5 (6.3)*** | 26.8 (8.8)*** | 10.2 (8.5)*** | 24.0 (8.8)*** |
| Length_next | 2.1 (6.5)* | 12.3 (8.2)*** | 3.7 (8.7)*** | 10.5 (8.5)*** | 3.4 (8.6)*** | 8.9 (8.8)*** |
| Frequency | 5.0 (4.7)*** | 29.4 (8.7)*** | 6.9 (4.6)*** | 65.2 (9.0)*** | 4.8 (6.1)*** | 81.3 (8.9)*** |
| Frequency_last | 1.8 (1.1) | 6.2 (8.8)*** | 1.6 (1.0) | 6.6 (8.6)*** | 6.6 (3.3)*** | 24.0 (8.8)*** |
| Frequency_next | 1.5 (5.5) | 17.5 (8.5)*** | 2.0 (7.5)* | 27.2 (8.7)*** | 2.7 (7.5)** | 22.6 (8.9)*** |
| **CCP** | 0.7 (1.0) | 12.1 (8.8)*** | 0.8 (3.0) | 17.4 (8.8)*** | 0.7 (1.0) | 37.1 (8.8)*** |
| CCP_last | 1.9 (6.0). | 2.1 (2.3) | **2.5 (6.0)*** | **5.5 (8.8)**** | 0.9 (2.0) | 14 (8.9)*** |
| CCP_next | 2.0 (1.0) | 19.0 (8.7)*** | **7.0 (1.0)**** | **14.1 (8.8)**** | 1.9 (1.0) | 8.5 (8.8)*** |
| **N-gram** | **2.7 (4.2)*** | **14.0 (6.4)**** | **3.8 (6.4)**** | **12.8 (8.9)**** | **4.2 (3.5)**** | **10.5 (8.8)**** |
| N-gram_last | **2.9 (7.4)**** | **12.3 (8.0)**** | **7.4 (1.0)**** | **14.6 (8.9)**** | 2.0 (5.6). | 18.6 (8.9)*** |
| N-gram_next | 3.7 (1.0). | 6.0 (8.4)*** | 2.7 (1.0) | 8.5 (7.7)*** | 5.3 (1.0)* | 10.7 (7.8)*** |
| **Topic** | 2.1 (2.4). | 14.6 (8.8)*** | 1.6 (2.9) | 14.2 (8.8)*** | 1.8 (2.1) | 13.4 (8.8)*** |
| Topic_last | 1.7 (5.4). | 19.6 (8.3)*** | 3.1 (6.5)** | 19.0 (8.4)*** | 1.9 (6.6). | 19.7 (8.5)*** |
| Topic_next | 1.8 (2.4) | 9.2 (8.6)*** | 0.3 (1.1) | 19.0 (8.4)*** | 0.2 (1.0) | 11.2 (8.9)*** |
| **RNN** | **4.4 (1.0)*** | **5.3 (8.5)**** | 1.5 (3.6) | 5.8 (8.1)*** | **8.6 (1.0)**** | **8.5 (8.4)**** |
| RNN_last | 1.5 (1.0) | 9.2 (8.8)*** | 3.3 (5.5)** | 5.7 (8.5)*** | 3.6 (3.9)** | 7.4 (8.8)*** |
| RNN_next | **2.3 (5.6)*** | **6.7 (7.8)**** | 1.6 (8.2) | 11.0 (7.4)*** | 1.8 (3.1) | 8.3 (8.7)*** |
| **Total $R^2$ (%)** | 5.45 | 5.36 | 9.16 | 11.5 | 10.7 | 11.3 |

. $p < 0.1$; * $p < 0.05$; ** $p < 0.01$; *** $p < 0.001$. Consistent findings over both data sets in **Tables 4–6** and this Table and are marked bold.

three viewing time measures. Moreover, the n-gram probability of the last word reproducibly accounted for variance in SFD and GD data. We illustrate these effects in **Figure 2**, suggesting that high log-transformed n-gram probabilities exhibit approximately linear decreases in GD, particularly in the range of log-transformed probabilities from −8 to −2 (cf. Smith and Levy, 2013).

The present and last-word n-gram effects can be consistently obtained in the analyses of only a single type of predictability (**Tables 4–6**), as well as in the analysis containing all predictors (**Table 8**). Moreover, the n-gram-based GAM, including the present, the last and the next word predictor, provided significantly better predictions than the CCP-based GAM in the SRC data set for SFD and TVT data (**Table 7**). This result pattern suggests that an n-gram model is more appropriate than CCP for predicting eye-movements in relatively small samples.

Concerning the non-linear analyses of the **topic** models, we found no effects that can be reproduced across all analyses (**Tables 4–6, 8**). Thus, we found an even less consistent picture for the topic model than for CCP. When tested against each other, CCP provided better predictions than the topics model in the GD data of the SRC, while the reverse result pattern was obtained for SFD data in the PSC (**Table 7**). In the analyses relying on a single type of predictability, we obtained a TVT effect for the present word that can be reproduced across both samples (**Table 6**). In the analyses containing all types of predictability, only the last-word's topical fit with the preceding sentence revealed a GD effect in the SRC data set that can be reproduced across both samples (**Table 8**). These result patterns may tentatively point at a late semantic integration effect of long-range semantics by topic model predictions.

The examination of **RNN** models revealed consistent findings across all non-linear analyses for the present word in SFD and TVT data (**Tables 4–6, 8**). Though consistent next-word predictability effects were obtained for all viewing time measures in the analyses containing only a single type of predictability (**Tables 4–6**), only the next-word RNN effect in SFD data was reproducible in the analyses containing all predictors (**Table 8**). This result pattern indicates that an RNN may be particularly useful for investigating (parafoveal) preprocessing in rapidly successful lexical access.

Therefore, we relied on SFD data to illustrate the functional form of the RNN effects in **Figure 3**. RNN probabilities of the present word reduce SFDs, particularly at a log-transformed probability of −10 and higher. Concerning the influence of the next word, log-probabilities lower than −7 seem to delay the SFDs of the current word. This provides some preliminary evidence that an extremely low RNN probability of the word in the right parafovea might leads to parafoveal preprocessing of the next word.

Next-word probability effects, however, are not the only domain, in which RNN models can account for other variance than the other types of predictability. We also obtained consistent last-word RNN-based GAM model fit improvements and significant effects of last-word probabilities in the analysis including all predictors for TVT data (**Tables 6**, **8**)—a result that probably points at the generalization capabilities of this subsymbolic model. For the SRC data set, the RNN provided significantly better predictions in SFD and TVT data, while CCP outperformed the RNN model in the TVT data of the PSC data set (**Table 7**).

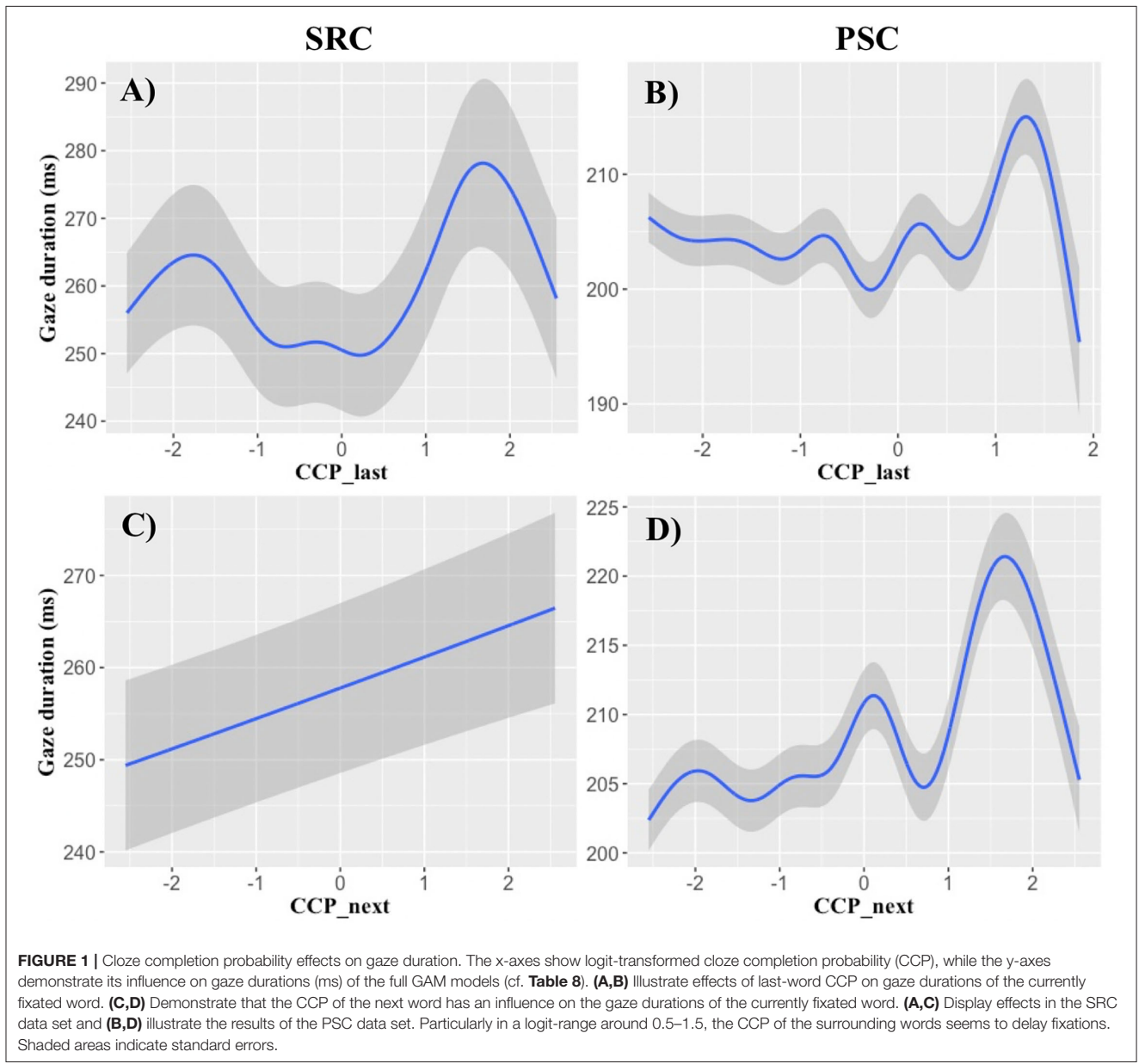

**FIGURE 1** | Cloze completion probability effects on gaze duration. The x-axes show logit-transformed cloze completion probability (CCP), while the y-axes demonstrate its influence on gaze durations (ms) of the full GAM models (cf. **Table 8**). **(A,B)** Illustrate effects of last-word CCP on gaze durations of the currently fixated word. **(C,D)** Demonstrate that the CCP of the next word has an influence on the gaze durations of the currently fixated word. **(A,C)** Display effects in the SRC data set and **(B,D)** illustrate the results of the PSC data set. Particularly in a logit-range around 0.5–1.5, the CCP of the surrounding words seems to delay fixations. Shaded areas indicate standard errors.

When summing up the results of language-model-based vs. the CCP-based GAM models, language models outperformed CCP in 5 comparisons, while CCP provided significantly better fitting GAM models in 2 comparisons (**Table 7**). Moreover, the three language models together always accounted for more viewing time variance than CCP (**Tables 4–6**). When the predictors of all three language models are together incorporated into a final GAM, this accounted for more viewing time variance than CCP (**Tables 4–6**). Thus, a combined set of language model predictors is appropriate to explain eye-movement patterns.

# GENERAL DISCUSSION

## Language Models Account for More Reproducible Variance

In the present study, we compared CCP to word probabilities calculated from n-gram-, topic- and RNN-models for predicting fixation durations in the SRC and PSC data set. Already simple item-level analyses showed that all language models provided greater linear correlations with all viewing time measures than CCP (**Table 3**). We also explored the possibility that the raw word probabilities provide a linear relationship with reading times (Brothers and Kuperberg, 2021). The transformed probabilities, however, always provided greater correlations with the three reading time measures than the raw probabilities. Therefore, our analyses support Smith and Levy's (2013) conclusion that the best prediction is obtained with log-transformed language model probabilities (cf. Wilcox et al., 2020).

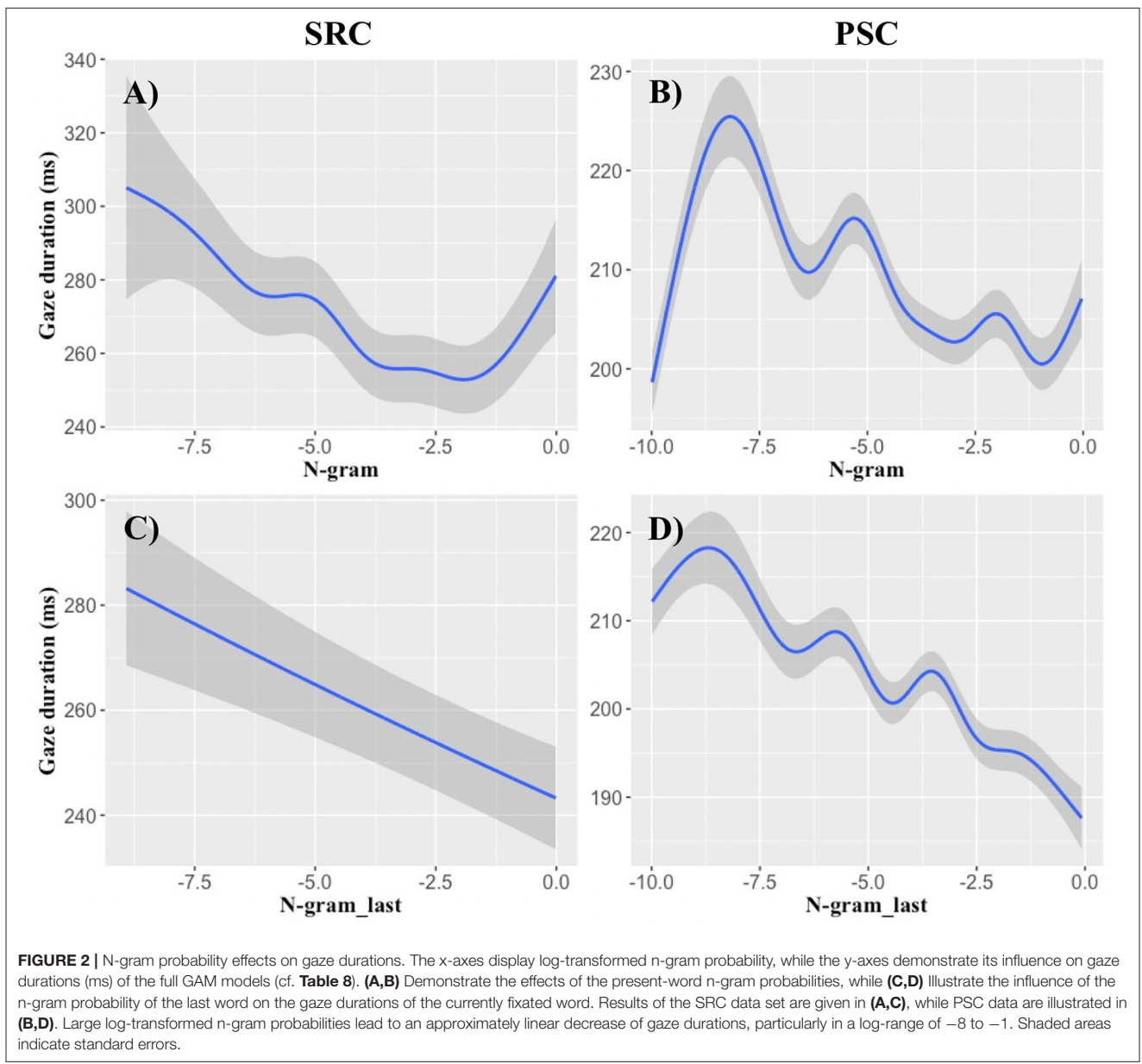

**FIGURE 2** | N-gram probability effects on gaze durations. The x-axes display log-transformed n-gram probability, while the y-axes demonstrate its influence on gaze durations (ms) of the full GAM models (cf. **Table 8**). **(A,B)** Demonstrate the effects of the present-word n-gram probabilities, while **(C,D)** Illustrate the influence of the n-gram probability of the last word on the gaze durations of the currently fixated word. Results of the SRC data set are given in **(A,C)**, while PSC data are illustrated in **(B,D)**. Large log-transformed n-gram probabilities lead to an approximately linear decrease of gaze durations, particularly in a log-range of −8 to −1. Shaded areas indicate standard errors.

When contrasting each language model against CCP in our non-linear analyses, there was no single language model that provided consistently better performance than CCP for the same viewing time measure in both data sets (**Table 7**). Rather, such comparisons seem to depend on a number of factors such as the chosen data set, language, participants and materials, demonstrating the need for further studies. A particularly important factor should be the number of participants in the CCP sample: In the small SRC data set, the language models outperformed CCP in 4 cases, while CCP was significantly better only 1 time. In the large CCP data set of the PSC, in contrast, both CCP and the language models outperformed each other for 1 time. When examining the amount of explained variance, however, the language models usually provided greater gains in explained variances than CCP: The n-gram and RNN models provided increments in explained variance ranging between 0.33 and 0.6% over CCP in the SFD and TVT data of the SRC data set, in which CCP however provided better predictions than the topic model (0.17%, **Table 7**). For the PSC data, there was a topic model gain of 0.1% over CCP in SFD data, but a CCP gain over the RNN model of 0.06% of variance. In sum, the language models provided better predictions than CCP in 5 cases—CCP provided better predictions in 2 cases (**Table 7**). Finally, the three language models together always outperformed CCP (**Tables 4–6**), supporting Hofmann's et al. (2017) conclusion derived from linear item-level based multiple regression analysis. Therefore, language models not only provide a deeper explanation for the consolidation mechanisms of the mental lexicon, but they also often perform better than CCP in accounting for viewing times.

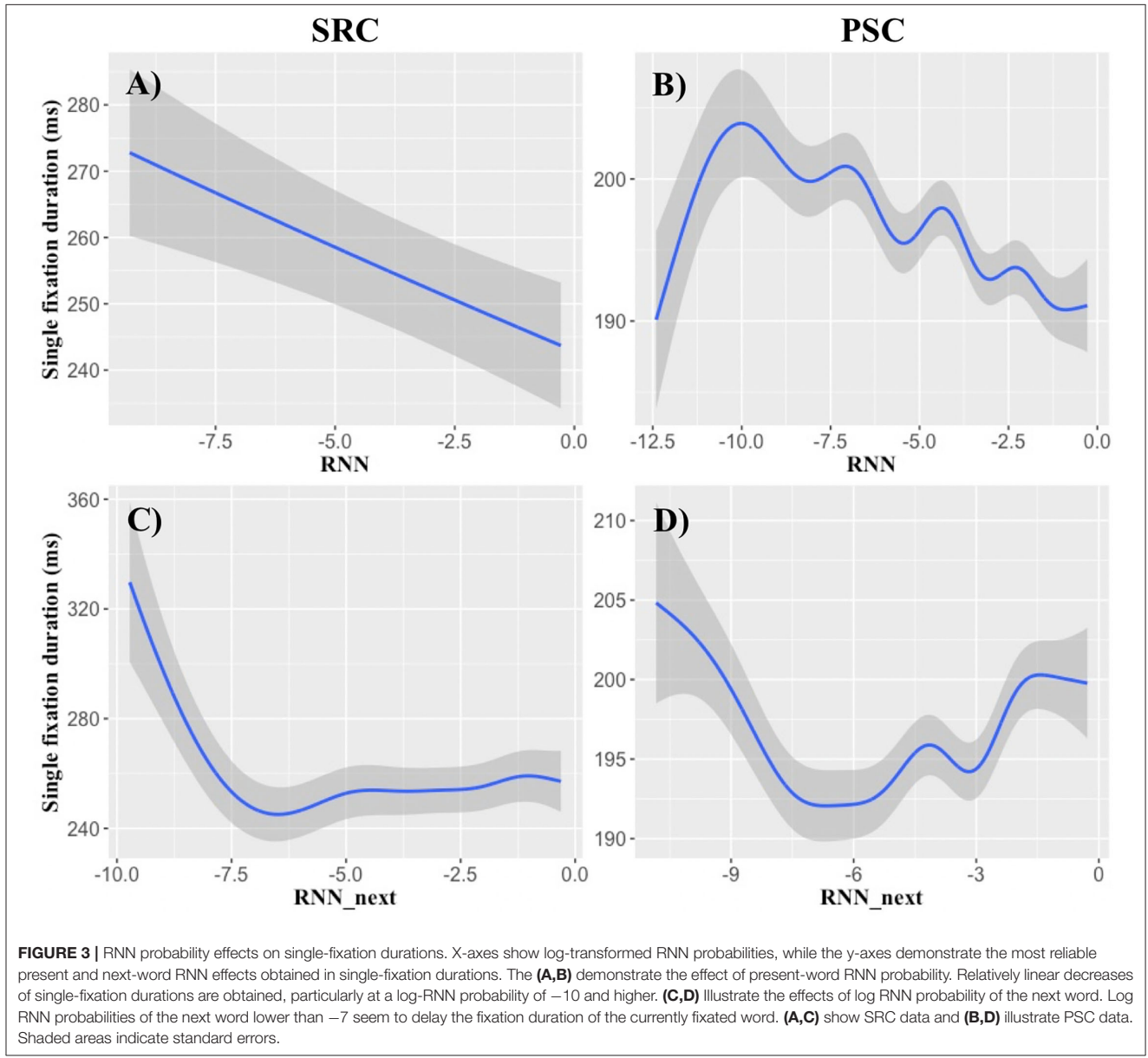

**FIGURE 3** | RNN probability effects on single-fixation durations. X-axes show log-transformed RNN probabilities, while the y-axes demonstrate the most reliable present and next-word RNN effects obtained in single-fixation durations. The **(A,B)** demonstrate the effect of present-word RNN probability. Relatively linear decreases of single-fixation durations are obtained, particularly at a log-RNN probability of −10 and higher. **(C,D)** Illustrate the effects of log RNN probability of the next word. Log RNN probabilities of the next word lower than −7 seem to delay the fixation duration of the currently fixated word. **(A,C)** show SRC data and **(B,D)** illustrate PSC data. Shaded areas indicate standard errors.

## CCP Effects Set a Challenge for Unexplained Predictive Processes

Nevertheless, CCP can still make a large contribution toward understanding the complex interplay of differential predictive processes. Though the results are less consistent than in the n-gram and RNN analyses, at least the last- and next-word CCP effects on GD data are reproducible in two eye movement samples and over several analyses (**Tables 5, 8**; cf. **Table 7**). This suggests that CCP accounts for variance that is not reflected by the language models we investigated (cf. e.g., Bianchi et al., 2020). When exploring the functional form of the GD effects of the surrounding words, **Figure 1** indicated that a high CCP of the last and next word leads to a relatively linear increase in GD, particularly in a logit-transformed predictability range of around 0.5–1.5. This might indicate that when between around 73 to 95% of the participants of the cloze completion sample agree that the cloze can be completed with this particular word, CCP might represent a reliable predictor of GD. As opposed to functional forms of the language model effects, CCP was the only variable that predicted an increase of viewing times with increasing levels of predictability. So CCP might represent an ongoing processing of the last word, or a type of parafoveal preprocessing that delays the present fixation (cf. **Figures 3C,D**, for other parafoveal preprocessing effects).

When having a large CCP sample available, the usefulness of CCP increases, as can be examined in the PSC effects of the present study (**Tables 4–8**). Though CCP can be considered an "all-in" variable, containing to some degree co-occurrence-based, semantic and syntactic responses (Staub et al., 2015), CCP samples might probably vary with respect to the amount of these different types of linguistic structure that is contained in these samples. This might in part explain some inconsistency between our SRC and PSC samples. We suggest that future studies should more closely evaluate which type of information is contained in which particular CCP sample, in order to obtain a scientifically deeper explanation for the portions of the variance that can presently still be better accounted for by CCP (Shaoul et al., 2014; Luke and Christianson, 2016; Hofmann et al., 2017; Lopukhina et al., 2021). **Table 3** shows that the correlations of n-gram and RNN models with the CCP data are larger than the correlations with topics models in both data samples. This suggests that the present cloze completion procedures were more sensitive to short-range syntax and semantics rather to long-range semantics. These CCP scores were based on sentence context—a stronger contribution of long-range semantics could be probably expected when the cloze (and reading) tasks are based on paragraph data (e.g., Kennedy et al., 2013).

Though last-word SFD effects (**Table 4**) and present-word TVT effects (**Table 6**) of CCP seem to be better explained by the language models (**Table 8**), this result pattern confirms the prediction of the EZ-reader model that CCP particularly reflects late processes during reading (Reichle et al., 2003).

## Symbolic Short-Range Semantics and Syntax in N-Gram Models

While it has often been claimed that CCP reflects semantic processes (e.g., Staub et al., 2015), it is difficult to define what "semantics" exactly means. Here we offer scientifically deep explanations relying on the computationally concrete definitions of probabilistic language models (Reichle et al., 2003; Westbury, 2016), which allow for a deeper understanding of the consolidation mechanisms. An n-gram model is a simple count-based model that relies exclusively on symbolic representations of words. We call it a short-range "semantics and syntax" model, because it is trained from the immediately preceding words. The n-gram model reflects sequential-syntagmatic "low-level" information (e.g., McDonald and Shillcock, 2003b).

The present word's n-gram probability accounted for early successful lexical access as reflected in SFD, standard lexical access as reflected in the GD, as well as late integration as reflected in the TVT. Moreover, the last word's n-gram probability accounted for lagged effects on SFD and GD data, which replicates and extends Smith and Levy's (2013) findings. The examination of the functional form also confirms their conclusion that last- and present-word log n-gram probability provides a (near-)linear relationship with GD (see also Wilcox et al., 2020)—at least in the log-transformed probability range of −8 to −2 (**Figure 2**). Such a near-linear relationship was also obtained for the log RNN probability of the present word with SFD data (**Figures 3A,B**).

The present- and last-word effects of the n-gram model were remarkably consistent across the two different eye-movement samples, as well as over different analyses (**Tables 4–6, 8**; cf. Wagenmakers et al., 2011). Our data support the view that n-gram models not only explain early effects of lexical access (e.g., McDonald and Shillcock, 2003a; Frisson et al., 2005), but can also be used for the study of late semantic integration (e.g., Boston et al., 2008). Moreover, they seem to be a highly useful tool, when aiming to demonstrate the limitations of the eye-mind assumption (Just and Carpenter, 1984). N-gram models consistently predict lagged effects that reflect the sustained semantic processing of the last word during the current fixation of a word. In sum, count-based syntactic and short-range semantic processes can reliably explain last-word and present-word probability effects (e.g., Smith and Levy, 2013; Baroni et al., 2014).

## Less Consistent Findings in Long-Range Topic Models

Topic models provided the least consistent picture over the two different samples and the different types of analyses (**Tables 4–6, 8**). Topic models are count-based subsymbolic approaches to long-range semantics, which is consolidated from the co-occurrences in whole documents. As the same is true for LSA, they can well be compared to previous LSA-findings (Kintsch and Mangalath, 2011). For instance, Bianchi et al. (2020) obtained remarkably weaker LSA-based effects than n-gram-based GD effects—thus our results are in line with the results of this study (cf. Hofmann et al., 2017). When they combined LSA with an n-gram model and additionally included next-word semantic matches in their regression model, some of the LSA-based effects became significant. This greater robustness of the LSA effects in Bianchi et al. (2020) may be well-explained by Kintsch and Mangalath's (2011) proposal, that syntactic factors should be additionally considered when aiming to address long-range meaning and comprehension. As opposed to Bianchi's et al. (2020) next-word GD effects, however, the present study revealed last-word GD effects of long-range semantics in the analysis considering all predictors (**Table 8**). And when examining only the present word's topical fit with the preceding sentence, Wang's et al. (2010) conclusion was corroborated that (long-range) lexical knowledge of whole documents is best reflected in TVT data (see **Table 6**).

In sum, long-range semantic models provide a less consistent picture than the other language models (cf. Lopukhina et al., 2021). The results become somewhat more consistent when short-range relations are additionally taken into account. Given that the effects occur in the last, present or next word when short-range semantics is included, this may point at a positional variability of long-range semantics that is integrated at some point in time during retrieval. We think that this results from the position-insensitive training procedure. Therefore, rather than completely rejecting the eye-mind assumption by proposing that "the process of retrieval is independent of eye movements" (Anderson et al., 2004, p. 225), we suggest that long-range semantics is position-insensitive at consolidation. Therefore, it is also position-insensitive at lexical retrieval and the effects can hardly be constrained to the last, present or next word, even if short-range relations are considered (Wang et al., 2010; Bianchi et al., 2020).

Finally, it should be taken into account that we here examined single sentence reading. This may explain the superiority of language models that are trained at the sentence level. Document-level training might be superior when examining words embedded in paragraphs or documents. This hypothesis is in part confirmed by Luke and Christianson (2016). They computed the similarity of each word to the preceding paragraph and found relatively robust LSA effects (Luke and Christianson, 2016, e.g., Tables 41–45).

## RNN Models: An Alternative View on the Mental Lexicon

Short-range semantics and syntax can be much more reliably constrained to present-, last-, or next-word processing, as already demonstrated by the consistent present and last-word n-gram effects. For the RNN probabilities, the simple linear item-level correlations with SFD, GD, and TVT data were largest, replicating, and extending the results of Hofmann et al. (2017). For the non-linear analyses, the present word's RNN probabilities provided the most consistent results for SFD and TVT (**Tables 5, 6, 8**). Though GD was significant when only examining a single language model (**Table 5**), this result could not be confirmed in the analysis in which all predictors competed for viewing time variance (**Table 8**). We propose that this difference can be explained by considering how these short-range models are trained for consolidated information. The n-gram model is symbolic; thus, the prediction is only made for a particular word form. And when testing for standard lexical access (that is reflected in the GD), a perfect match of the predicted and the observed word form may explain the superiority of the n-gram model in this endeavor (**Table 8**).

On the other hand, both language models accounted for SFD and TVT variance in the analysis containing all types of predictability (**Table 8**). The co-existence of these effects may be explained by the proposal that both models represent different views on the mental lexicon (Elman, 2004; Frank, 2009). The n-gram model represents a "static" view, in which large lists of word sequences and their frequencies are stored, to which word is more likely to occur in a context, given this large "dictionary" of word sequences. The RNN model, in contrast, has only a few hundred hidden units that reflect the "mental state" of the model (Elman, 2004). As a result of this small "mental space," neural models have to compress the word information, which may, for instance, explain their generalization capabilities: When such a model is trained to learn statements such as "robin is a bird" and "robin can fly," and it later learns only a few facts about a novel bird, e.g., "sparrow can fly," "sparrow" obtains a similar hidden unit representation as "robin" (McClelland and Rogers, 2003). Therefore, a neural model can complete the cloze "sparrow is a …" with "bird," even if it never was presented this particular combination of words. For instance, in our PSC example stimulus sentence "Die [The] Richter [judges] der [of the] Landwirtschaftsschau [agricultural show] prämieren [award a price to] Rhabarber [rhubarb] und [and] Mangold [mangold]," the word "prämieren" (award a price) has a relatively low n-gram probability of 1.549e-10, but a higher RNN probability of 1.378e-5, because the n-gram model has never seen the word n-gram "der Landwirtschaftsschau prämieren [agricultural show award a price to]," but the RNN model's hidden units are capable of inferring such information from similar contexts (e.g., Wu et al., 2021). In sum, both views on the mental lexicon account for different portions of variance in word viewing times (**Table 8**). The n-gram model may explain variance resulting from an exact match of a word form in that context, while for instance generalized information may be better explained by the RNN model.

There are, however, also differences in the result patterns that are best captured by the two language models, respectively. A notable difference between count-based, symbolic knowledge in the n-gram vs. predict-based, subsymbolic knowledge in the RNN lies in their capability to account for last-word vs. next-word effects. While the n-gram model obtained remarkably consistent findings for the last word, next-word SFD effects are better captured by an RNN (**Tables 4, 8**). This corroborates our conclusion that the two views on the mental lexicon account for different effects. The search for a concrete word form, given the preceding word forms in the n-gram model, may take some time. Therefore, the probabilities of the last word still affect the processing of the present word. As opposed to this static view on a huge mental lexicon, the hidden units in the RNN model are trained to predict the next word (Baroni et al., 2014). When such a predicted word to the right of the present fixation position may cross a log RNN probability of −7, its presence can be verified (**Figures 3C,D**). Therefore, the RNN-based probability of the next word may elicit fast and successful lexical access of the present word, as reflected in the SFD.

## Limitations and Outlook

Model comparison is known to be related to the numbers of parameters included in the models, thus a comparison of the GAMs comparing all three language models with the CCP predictor might overemphasize the effects of the language models (**Tables 4–6**). However, we think that language models allow for a deeper understanding of natural language processing of humans than CCP does, because language models provide a computational definition how "predictability" is consolidated from experience. Moreover, the conclusion that language models account for more variance than CCP is also corroborated by all other analyses (**Tables 3–8**). Sometimes it is stated that GAMs in general are prone to overfitting. CCP and even more so the model-generated predictability scores are highly correlated with word frequency, potentially leading to overfitting on the one hand. On the other hand, it is more difficult for the language models to account for additional variance, because their higher correlation (i.e., their higher shared variances) with word frequency (**Table 3**). Wood (2017) discusses that the penalization procedure that leads to the smoothing parameters and also the GCV procedure inherent in the gam() function in R specifically tackles overfitting. We further addressed this question by focusing on consistent results, visible across the computation for two independent eye-movement samples and different types of analyses. This approach reduced the number of robust and consistent findings. If one would have ignored the non-linear nature of the relationships between predictors and dependent variables and only examined the simple linear effects reported in **Table 3**, however, the advantages of the language models over CPP would have been much clearer.

This also leads to the typical concern for analyses on unaggregated data that they account for only a small "portion" of variance. For instance, Duncan (1975, p. 65) suggests to "eschew altogether the task of dividing up $R^2$ into unique causal components" (quoted from Kliegl et al., 2006, p. 22). It is clear that unaggregated data contain a lot of noise, for instance resulting from a random walk in the activation of lexical units, random errors in saccade length, and also "[s]accade programs are generated autonomously, so that fixation durations are basically realizations of a random variable" (Engbert et al., 2005, p. 781). Therefore, if we like to estimate the "unique casual component" of semantic and syntactic processes, we need to rely on aggregated data. **Table 3** suggests that these top-down factors represented by language account for a reasonable 15–36% of the viewing time variance, while CCP accounted for 7–19%.

It also becomes clear that the two data sets (PSC and SRC), differ in important aspects. First, the CCP samples differ in size, thus they probably provide a different signal-to-noise ratio. A second difference is that also the PSC eye-movement sample is larger and thus has more statistical power to identify significant eye-movement effects of single predictors. Third, the CCP measure is derived from different participant samples. We would also point to the fact, that the eye-movement and CCP samples were collected at different times and come from different countries, which may also explain some differences between the obtained effects. Fourth, also the English and the German subtitles training corpora may contain slightly different information. Having these limitations in mind, we feel that the most consistent findings discussed in the previous sections represent robust effects to evaluate the functioning and the predictions of language models. Our rather conservative approach might miss some effects that are actually apparent, though they are explainable by these four major differences between the samples. Therefore, future studies might more closely characterize the CCP participant samples. They may examine the same participants' eye movements to obtain predictability estimates that are representative for the participants—which might increase the amount of explainable variance (cf. Hofmann et al., 2020). Finally, google n-gram training corpora may help to obtain training corpora stemming from the same time as the eye-movement data collection.

Though we were able to discriminate between long-range semantics and short-range relations that can be differentiated into count-based symbolic and predict-based subsymbolic representations, we like to point at the fact that the short-range relations could also be separated into semantic and syntactic effects. For example, RNN models have previously also been related to syntax processing (Elman, 1990). Therefore, syntactic information may alternatively explain the very early and very late effects (Friederici, 2002). To examine whether semantic or syntactic effects are at play, a promising start for further evaluations may examine content vs. function words, which may even lead to more consistent findings for long-range semantic models. Further analysis may focus on language models that take into account syntactic information (e.g., Padó and Lapata, 2007; cf. Frank, 2009).

## CONCLUSION

Understanding the complex interplay of different types of predictability for reading is a challenging endeavor, but we think that our review and our data point at differential contributions of count-based and predict-based models in the domain of short-range knowledge. Count-based models better capture last-word effects, predict-based models better capture early next-word effects, while present-word probabilities both make an independent contribution to viewing times. In contrast, CCP is a rather all-in predictor, that probably covers both types of semantics: short-range and long-range. But we have shown that language models with their differential foci are better suited for a deeper explanation for eye-movement behavior, and thus applicable in theory development for models of eye-movement control. Finally, we hope that we made clear that these relatively simple language models are highly useful for understanding differential lexical access and semantic integration parameters that are reflected in differential viewing time parameters.

## DATA AVAILABILITY STATEMENT

Publicly available datasets were analyzed in this study. These data can be found at: https://clarinoai.informatik.uni-leipzig.de/fedora/objects/mrr:11022000000001F2FB/datastreams/EngelmannVasishthEngbertKliegl2013_1.0/content. The data and analysis examples of the present study can be found under https://osf.io/z7d3y/?view_only=be48ab71ccd14da5b0413269c150d2f9.

## ETHICS STATEMENT

Ethical review and approval was not required for the study on human participants in accordance with the local legislation and institutional requirements. Written informed consent for participation was not required for this study in accordance with the national legislation and the institutional requirements.

## AUTHOR CONTRIBUTIONS

SR provided the language models. MH analyzed the data. LK checked and refined the analyses. All authors wrote the paper (major writing: MH).

## FUNDING

This paper was funded by a grant of the Deutsche Forschungsgemeinschaft to MH (HO 5139/2-1 and 2-2).

## ACKNOWLEDGMENTS

We like to thank Albrecht Inhoff, Arthur Jacobs, Reinhold Kliegl, and the reviewers of previous submissions for their helpful comments.

**Conflict of Interest:** The authors declare that the research was conducted in the absence of any commercial or financial relationships that could be construed as a potential conflict of interest.

**Publisher's Note:** All claims expressed in this article are solely those of the authors and do not necessarily represent those of their affiliated organizations, or those of the publisher, the editors and the reviewers. Any product that may be evaluated in this article, or claim that may be made by its manufacturer, is not guaranteed or endorsed by the publisher.